\documentclass[11pt]{article}

\usepackage[preprint]{acl}

\usepackage{times}
\usepackage{latexsym}

\usepackage[T1]{fontenc}

\usepackage[utf8]{inputenc}

\usepackage{microtype}

\usepackage{inconsolata}

\usepackage{graphicx}

\usepackage{amsmath}
\usepackage{amssymb}

\usepackage{booktabs}

\usepackage{float}
\usepackage{placeins}

\usepackage{xcolor}

\title{Weight Tying Biases Token Embeddings Towards the Output Space}

\author{
 \textbf{Antonio Lopardo\textsuperscript{1}},
 \textbf{Avyukth Harish\textsuperscript{2}},
 \textbf{Catherine Arnett\textsuperscript{1}},
 \textbf{Akshat Gupta\textsuperscript{2}}
\\
 \textsuperscript{1}EleutherAI,
 \textsuperscript{2}University of California Berkeley
\\
 \small{
   \textbf{Correspondence:} \href{mailto:antonio.lopardo@outlook.com}{antonio.lopardo@outlook.com}
 }
}

\begin{document}
\maketitle
\begin{abstract}
Weight tying, i.e. sharing parameters between input and output embedding matrices, is common practice in language model design, yet its impact on the learned embedding space remains poorly understood. 
In this paper, we show that tied embedding matrices align more closely with output (unembedding) matrices than with input embeddings of comparable untied models, indicating that the shared matrix is shaped primarily for output prediction rather than input representation.
This unembedding bias arises because output gradients dominate early in training. Using tuned lens analysis, we show this negatively affects early-layer computations, which contribute less effectively to the residual stream.
Scaling input gradients during training reduces this bias, providing causal evidence for the role of gradient imbalance. 
This is mechanistic evidence that weight tying optimizes the embedding matrix for output prediction, compromising its role in input representation. 
These results help explain why weight tying can harm performance at scale and have implications for training smaller LLMs, where the embedding matrix contributes substantially to total parameter count.
\end{abstract}

\section{Introduction}
\label{sec:introduction}

Weight tying has become standard practice in language model design since \citet{press2017using} showed that sharing parameters between the input embedding matrix and the output projection matrix improves performance while reducing parameter count in recurrent models. This practice is often motivated by parameter efficiency, implicit regularization, and the intuitive symmetry between reading and predicting tokens.

\begin{figure}[ht]
    \centering
    \includegraphics[width=\columnwidth]{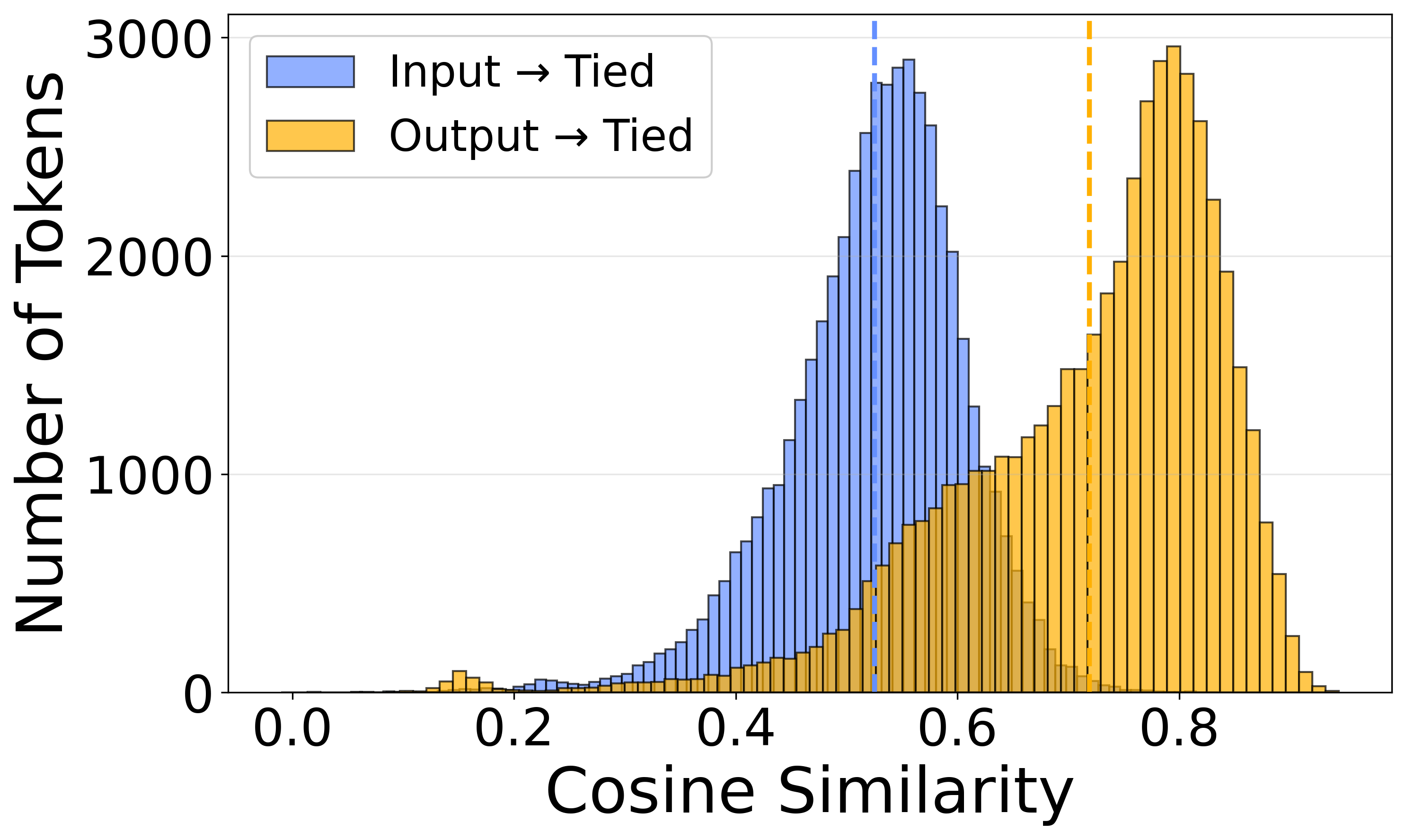}
    \caption{Per-token cosine similarity (after linear alignment) between the tied embedding matrix and the untied input (blue) and output (orange) matrices for two OLMo-1B training runs (tied and untied). Dashed lines indicate means. The tied matrix is substantially more aligned with the untied output (mean: 0.719) than the untied input (mean: 0.525).
    }
    \label{fig:token-alignment}
\end{figure}

However, more recent work has challenged the applicability of these findings to modern architectures. \citet{chung2020rethinking} showed that weight tying can be harmful to performance in deeper transformer-based language models \citep{vaswani2017attention}, where the benefits observed in shallow networks no longer hold.

This finding is reflected in contemporary model design choices: the Qwen3 family \citep{qwen3technicalreport} explicitly ties embeddings for smaller models but unties them for larger ones, suggesting that the costs of weight tying outweigh its benefits as models grow. Similarly, OLMo 2 and 3 \citep{olmo2furious, olmo3} explicitly untie embeddings for all model sizes.
Beyond these architectural trends, \citet{machina2024anisotropy} found evidence that Pythia models, which use untied embeddings, exhibit less anisotropy in their representation spaces compared to tied models, suggesting that untying may yield more uniform representations, a property shown to improve downstream performance \citep{gao2019representation, ethayarajh2019contextual, bis2021toomuch, rajaee2022isotropy, wang2019adversarial, wang2020spectrum, stollenwerk2025coupled}. 

These observations collectively suggest untying is beneficial at scale. However, the underlying mechanism behind why this is the case remains poorly understood. In this paper, we provide an explanation to this exact question, grounded in training dynamics. We show that in weight-tied models, the shared matrix is dominated early in training by gradients from the output layer and thus receives stronger learning signals from the next-token prediction objective.
As a consequence, \textbf{the shared embedding matrix is shaped primarily for output projection, leaving early transformer layers to operate with input embeddings optimized for a different purpose} (Figure~\ref{fig:token-alignment}).
We support these findings with embedding similarity analysis, analysis with the tuned lens \citep{belrose2023tunedlens}, gradient flow tracking, and gradient manipulation experiments. Together, these experiments show that the tied embedding matrix is largely shaped by its role in the output layer, due to gradient imbalances early in training.\footnote{Code and artifacts: \url{https://github.com/AntonioLopardo/weight-tying-bias}}

In summary:
\begin{itemize}
    \item We show that tied embedding matrices align more closely with output (unembedding) matrices than with input embeddings for comparable untied models.
    \item This unembedding bias affects early-layer computations, as shown by elevated KL divergence in tuned lens analysis.
    \item We show that output-layer gradients dominate input-layer gradients early in training, resulting in the unembedding bias in weight-tied models.
    \item Finally, we show that scaling input gradients during training reduces this bias, confirming the causal role of gradient imbalance.
\end{itemize}

\section{Background \& Related Work}
\label{sec:background}\label{subsec:input-output-roles}\label{sec:related-work}


In transformer-based language models, the input embedding matrix $W_E \in \mathbb{R}^{V \times d}$ (where $V$ is the vocabulary size and $d$ is the hidden dimension of the model) maps discrete token indices to continuous vector representations that serve as input to the first transformer layer. These initial representations encodes semantic and syntactic information in a form that subsequent layers can process. The unembedding matrix $W_U \in \mathbb{R}^{d \times V}$ serves a different function: it projects the final hidden states back to vocabulary-sized logits for next-token prediction. This matrix must transform the refined representations from the last layer into logits over the vocabulary, where each logit reflects the likelihood of the corresponding token given the context.

The input embedding matrix should thus produce representations useful for the model's internal computations, while the output matrix must discriminate between tokens based on the final hidden state, optimizing directly for cross-entropy loss. With tied embeddings, a single set of parameters must satisfy both objectives simultaneously \citep{bertolotti2024tying}. We show that in this scenario one objective dominates: that of the output.

\subsection{Weight Tying in Language Models}
\label{subsec:weight-tying-lms}

Weight tying for embedding and unembedding matrices was introduced by \citet{press2017using} and \citet{inan2017tying}, who showed that sharing the input embedding matrix with the unembedding matrix, such that $W_U = W_E^\top$, improves perplexity while reducing parameter count in Recurrent Neural Networks \citep[RNNs;][]{elman1990finding}. The technique became standard practice in transformer language models, adopted by GPT-2 \citep{radford2019language}, BERT \citep{devlin2019bert}, RoBERTa \citep{liu2019roberta}, and Gemma \citep{team2024gemma}, \textit{inter alia}.

The parameter savings from weight tying are substantial, particularly for smaller models. For example, for Pythia 70M, the embedding matrices account for over 73.4\% of all parameters; even at 2.8B scale, they still represent approximately 9.2\% of the model, so tying can still yield meaningful savings. See Appendix \ref{app:param_counts} for elaboration.
However, \citet{chung2020rethinking} demonstrated that weight tying can harm performance: decoupling input and output embeddings improves average performance on the XTREME benchmark (\citealp{hu2020xtreme}; more details in Appendix \ref{app:appendix-chung}). This shows models benefit from specialized input and output embedding spaces. 
Currently, untying is more common for models at multi-billion parameters scale, e.g. Llama~3 \citep{grattafiori2024llama3}, OLMo~2 \citep{olmo2furious}, DeepSeek-V3 \citep{deepseekai2024deepseekv3}, and Qwen3 \citep{qwen3technicalreport}. In this work, we aim to understand the effect of weight tying on learned representations, as this becomes a more common design choice. 

\subsection{Embedding Space Alignment}
\label{subsec:embedding-space-alignment}

The alignment of embedding spaces has a long history in NLP, starting with work on cross-lingual word embeddings. \citet{mikolov2013exploiting} first showed that crosslingual embedding spaces share similar geometric structure, enabling alignment via simple linear transformations. \citet{xing2015normalized} demonstrated that constraining this mapping to be orthogonal significantly improves alignment quality, and \citet{artetxe2016learning} further refined this approach with principled bilingual mappings that preserve monolingual structure. \citet{conneau2020emerging} applied these techniques to contextual representations, demonstrating that even independently trained monolingual BERT models can be aligned post-hoc with linear mappings. We adopt a similar approach, using different transformations, to measure alignment difficulty between matrices across tied and untied models.

\subsection{The Logit Lens and the Tuned Lens}
\label{subsec:logit-tuned-lens}

The logit lens, introduced by \citet{nostalgebraist2020logitlens}, projects hidden states from intermediate layers into vocabulary distributions by applying the model's unembedding matrix $W_U$. For a model with $L$ total layers, the logit lens operator transforms an intermediate layer representation $h_\ell \in \mathbb{R}^d$ ($\ell < L$) as shown below:
\[
\mathrm{LogitLens}(h_\ell) = \mathrm{LayerNorm}(h_\ell) \, W_U
\]
This technique revealed that transformer predictions converge roughly monotonically across layers as the residual stream accumulates updates from each layer. However, the logit lens assumes intermediate representations are in the same basis as the final layer, which is often not the case. The tuned lens \citep{belrose2023tunedlens} addresses this limitation by learning an affine transformation for each layer:
\[
\mathrm{TunedLens}_\ell(h_\ell) = \mathrm{LogitLens}(A_\ell h_\ell + b_\ell)
\]
where $A_\ell \in \mathbb{R}^{d \times d}$ is a learned linear map and $b_\ell \in \mathbb{R}^d$ is a bias term. The translators compensate for cross-layer representational differences. The difficulty of learning these translators, measured by the residual KL divergence between the tuned lens prediction and the model's final output distribution, indicates how compatible each layer's representations are with the final layer. In our experiments, therefore, we use the tuned lens over logit lens.

\section{Models}
\label{subsec:models}

Our analysis uses open-weights models with both tied and untied configurations at the billion-parameter scale.

\paragraph{GPT-Neo and Pythia} GPT-Neo \citep{black2022gptneox} is an open-source replication of GPT-3, using tied embeddings for smaller sizes. The Pythia suite builds on the GPT-Neo architecture but unties embeddings for all models, comprising eight model sizes from 70M to 12B parameters, with 154 released checkpoints each. Both model families are trained on the Pile \citep{gao2020pile} and share the same codebase and training infrastructure, lending themselves to comparison. We use the 1B and 2.8B models.

\paragraph{OLMo-1B} The Open Language Model \citep{groeneveld2024olmo} provides fully open-source models with publicly released training code, data, and intermediate checkpoints. Crucially, the developers released two independent OLMo-1B training runs: \texttt{OLMo-1B}\footnote{\url{https://huggingface.co/allenai/OLMo-1B-hf}} (tied, February 2024) and \texttt{OLMo-1B-0724}\footnote{\url{https://huggingface.co/allenai/OLMo-1B-0724-hf}} (untied, July 2024). Both share the same core architecture but were trained on different versions of the Dolma dataset \citep{soldaini2024dolma}, making them good candidates for isolating the effect of weight tying on training dynamics.

\paragraph{Qwen3} The Qwen3 family uses scale-dependent weight tying: smaller models (0.6B, 1.7B, and 4B) tie embeddings while larger models (8B and above) untie them. All models share training methodology, tokenizer, and architectural design principles, varying only in parameter count\footnote{We note that unlike GPT-Neo, Pythia, and OLMo, training data are closed and may not be the same across models.}. We compare Qwen3-4B (tied) with Qwen3-8B (untied) to replicate our OLMo findings across different model families. While model size is a confound, the shared training setup helps isolate the effect of weight tying.


\section{Embedding Space Alignment Analysis}
\label{sec:alignment-analysis}

\subsection{Representational Alignment}

\label{subsec:alignment-methodology}

To test whether tied embeddings more closely resemble input or output representations, we compare the tied embedding matrix to both input and output embedding matrices from corresponding untied models. Following prior work \citep{conneau2020emerging}, we learn transformations to align embeddings between matrix pairs and evaluate alignment quality using mean cosine similarity between corresponding token vectors after alignment.

We consider three transformation types of increasing expressiveness:
\begin{itemize}
    \item \textbf{Identity}: Without transformations, we measure whether embeddings are already aligned. 
    \item \textbf{Orthogonal}: With a  Procrustes analysis \citep{schonemann1966}, which preserves distances and angles, we test whether the spaces differ only by a rotation transformation.
    \item \textbf{Linear}: Using an unconstrained linear mapping via least squares and allowing arbitrary rescaling, we ask if a linear relationship exists between the spaces.
\end{itemize}

Table~\ref{tab:alignment-results} shows the alignment results for OLMo-1B and GPT-Neo/Pythia. Because Qwen models have different embedding dimensions, we instead use KNN overlap and spectral distance analysis (Appendix~\ref{app:appendix-knn}).

For each transformation, we align embedding matrices pairwise: \textit{Input (U) $\rightarrow$ Output (U)} aligns the untied input and output matrices, \textit{Output (U) $\rightarrow$ Tied} aligns the untied output to the tied matrix, and \textit{Input (U) $\rightarrow$ Tied} aligns the untied input to the tied matrix. We compare OLMo-1B (tied) against OLMo-1B-0724 (untied), and GPT-Neo-2.7B (tied) against Pythia-2.8B (untied).
If output-layer gradients dominate the shared matrix, we expect that aligning the untied output matrix to the tied matrix should yield higher similarity than aligning the untied input matrix to the tied matrix. We find:

\paragraph{The tied embedding matrix is more aligned with the output embedding matrix of corresponding untied models than the input embedding matrix.} For both OLMo and Pythia models, the cosine similarity score for \textit{Output (U) $\rightarrow$ Tied} is much larger than the similarity scores for \textit{Input (U) $\rightarrow$ Tied} for all three transformations (Table~\ref{tab:alignment-results}). This shows that the shared embedding matrix in tied models resemble what an untied output matrix would look like much more than an untied input matrix. Supplemental KNN overlap and spectral distance analyses corroborate this finding across all three model families (see Appendix~\ref{app:appendix-knn}).

\begin{table}[ht]
\centering
\small
\setlength{\tabcolsep}{3pt}
\begin{tabular}{lccc}
\toprule
\textbf{Comparison} & 
\textbf{Identity} & 
\textbf{Orthogonal} & 
\textbf{Linear} \\
\midrule
\multicolumn{4}{l}{\textit{OLMo-1B (tied) vs OLMo-1B-0724 (untied)}} \\
Input (U) $\rightarrow$ Output (U) & 0.012 & 0.440 & 0.574 \\
Output (U) $\rightarrow$ Tied & \textbf{0.014} & \textbf{0.669} & \textbf{0.719} \\
Input (U) $\rightarrow$ Tied & 0.001 & 0.420 & 0.525 \\
\midrule
\multicolumn{4}{l}{\textit{GPT-Neo-2.7B (tied) vs Pythia-2.8B (untied)}} \\
Input (U) $\rightarrow$ Output (U) & -0.001 & 0.456 & 0.518 \\
Output (U) $\rightarrow$ Tied & \textbf{0.001} & \textbf{0.507} & \textbf{0.637} \\
Input (U) $\rightarrow$ Tied & 0.000 & 0.376 & 0.507 \\
\bottomrule
\end{tabular}
\caption{Geometric alignment between embedding spaces under identity, orthogonal (rotation), and linear (unconstrained least squares) transformations. We report mean cosine similarity between corresponding token embeddings from tied and untied models.}
\label{tab:alignment-results}
\end{table}

\paragraph{Untied input and output matrices exhibit limited similarity} The alignment between untied embeddings yields only moderate similarity even after learning a linear transformation between the two spaces (0.574 for OLMo, 0.440 for GPT-Neo/Pythia; \textit{Input (U) $\rightarrow$ Output (U)} rows in Table~\ref{tab:alignment-results}). This  suggests that input and output embedding representations do not converge when embeddings are untied. Thus, we argue that tying forces a compromise between representational demands.

\subsection{Tuned Lens Reveals ``First Layer Penalty"}
\label{subsec:tuned-lens-analysis}

The alignment analysis above reveals that tied embeddings structurally resemble output embeddings. But does this structural similarity have mechanistic consequences for how the model processes information? To investigate, we turn to the tuned lens.

The tuned lens (Section~\ref{subsec:logit-tuned-lens}) trains an affine translator at each layer to minimize the KL divergence between the translated hidden state's prediction distribution and the model's final output distribution. The residual KL divergence after training indicates how well each layer's representations align with the output space, lower values suggest better alignment, making this metric particularly informative for our analysis. 

\paragraph{Method} We specifically focus on the residual KL-divergence after training translators in early layers. Because weight tying makes the input and output embedding matrices identical, one might expect reduced input–output mismatch and therefore lower tuned-lens KL divergence at early layers. On the contrary, if weight tying forces the embedding matrix to prioritize output prediction, we instead can expect early layers in tied models to produce representations that are \emph{less} compatible with the output space, since these early layers must work with input embeddings optimized for a different purpose. This would manifest as higher KL divergence at early layers compared to untied models. 

\begin{figure}[ht]
    \centering
    \includegraphics[width=\columnwidth]{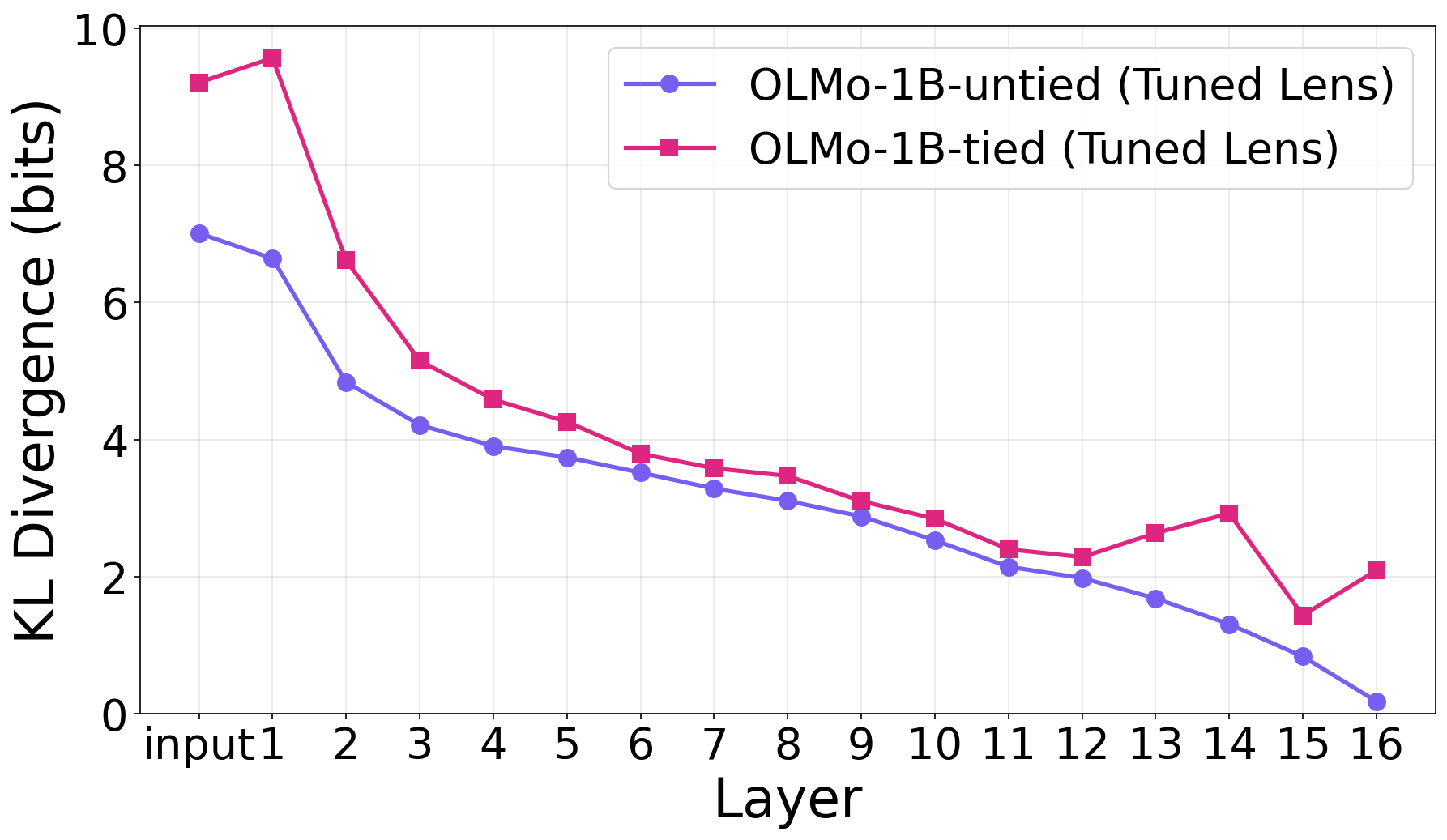}
    \caption{Tuned lens KL divergence for tied vs untied OLMo-1B. Lower values indicate better alignment between a layer's representations and the output space. The clearest separation appears in the early layers, where the tied model shows higher KL divergence.}
    \label{fig:olmo1b-comparison}
\end{figure}

\paragraph{Results} The results for residual KL-divergence after training translators, for the same number of steps, for OLMo-1B (tied vs untied) are shown in Figure~\ref{fig:olmo1b-comparison}. We additionally report results for Pythia-2.8B vs GPT-Neo-2.7B and Qwen3-4B vs Qwen3-8B in Appendix~\ref{app:appendix-tuned-lens}. We see that \textbf{tied models exhibit systematically higher KL divergence than untied counterparts in the early layers.} For OLMo-1B, the tied model starts at approximately 9.2 bits at the input layer compared to 7.0 bits for the untied model, a gap of over 2 bits. This gap persists across the early layers before gradually narrowing, indicating that the first few layers of the tied model operate with representations that are harder to align with the output space.

This elevated divergence suggests that early-layer activations in tied models are harder to align with the output space. The first transformer layer, which operates on representations produced by the tied embedding matrix, contributes less effectively to the residual stream. Later layers must compensate for this initial mismatch rather than building predictions incrementally from the start.

\section{Output Gradient Dominance}
\label{sec:output-dominance}

Our alignment analysis reveals a clear unembedding bias: tied embeddings more closely resemble untied output embeddings than untied input embeddings.  And our tuned lens analysis shows the implications of this mismatch, where the early layers in tied models contribute less effectively to the residual stream. In this section, we explore the reason behind the dominance of the tied output embeddings in learning the shared embedding vectors by examining gradient flow in both untied and tied models. 


\begin{figure}[t]
    \centering
    \makebox[\columnwidth][c]{\includegraphics[width=1.0\columnwidth]{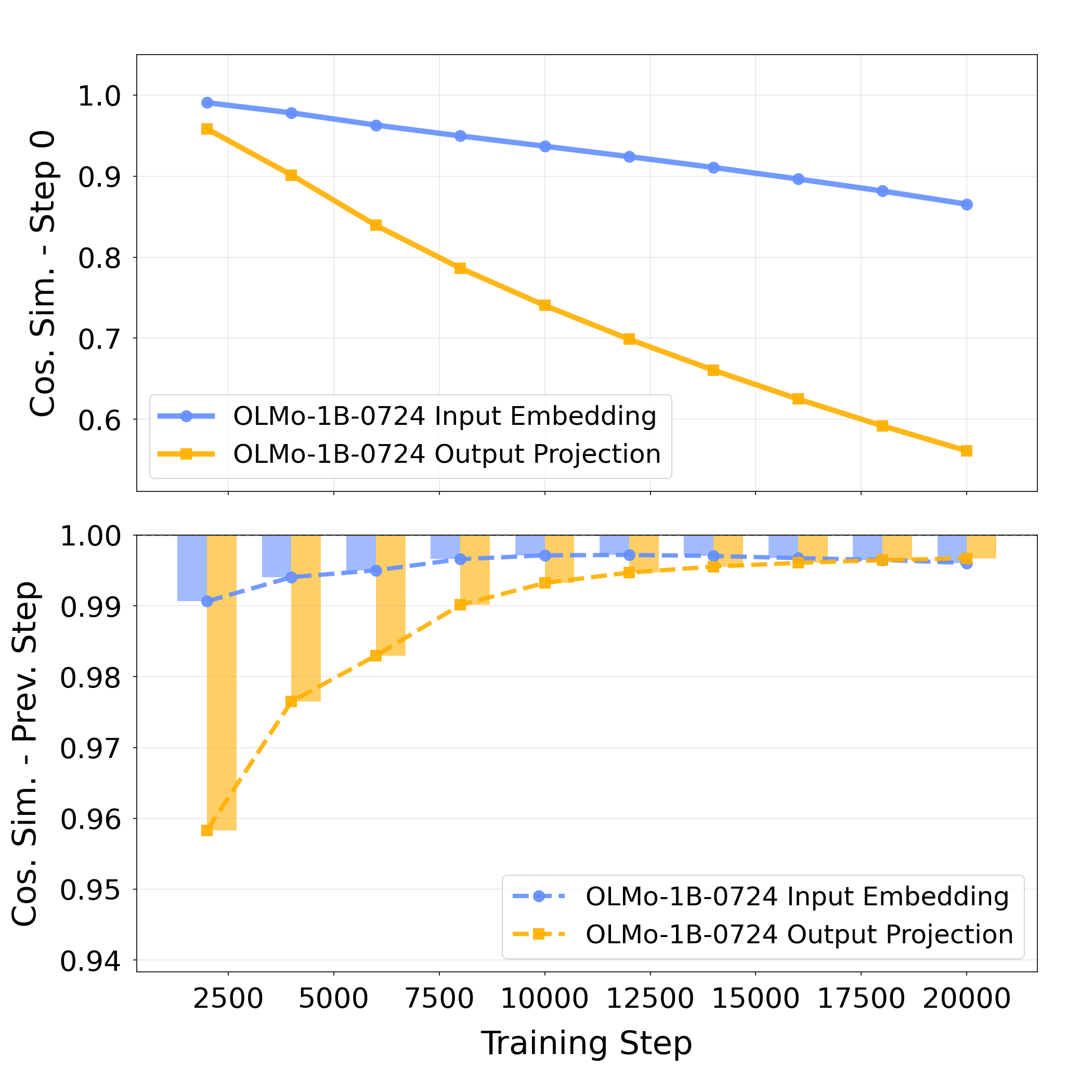}}
    \caption{For OLMo-1B-0724 (untied). Top: cosine similarity to initial embeddings (step 0), measuring cumulative drift. Bottom: cosine similarity between consecutive checkpoints, measuring between checkpoints change rate. 
    }
    \label{fig:embedding-changes-olmo}
\end{figure}

\subsection{Output-Embedding Evolve Faster than Input-Embeddings in Untied Models}
\label{subsec:dominance-methodology}

In untied models, the input and output embedding matrices are separate and receive gradients independently. Due to the direct connection between the output layer and the cross-entropy loss, the output embedding matrix may change more rapidly during training. By tracking how each matrix evolves across model checkpoints, we can establish whether this asymmetry exists even when the matrices are not tied.

This analysis requires access to model weights at multiple points during training. \textbf{Pythia} \citep{biderman2023pythia} released many checkpoints for all models, saved at regular intervals during training; \textbf{OLMo} \citep{groeneveld2024olmo} releases checkpoints every 1000 training steps, too. Both families have models with untied embeddings, allowing us to track input and output matrices independently.

\begin{figure*}[ht]
    \centering
    \makebox[\textwidth][c]{\includegraphics[width=1.05\textwidth]{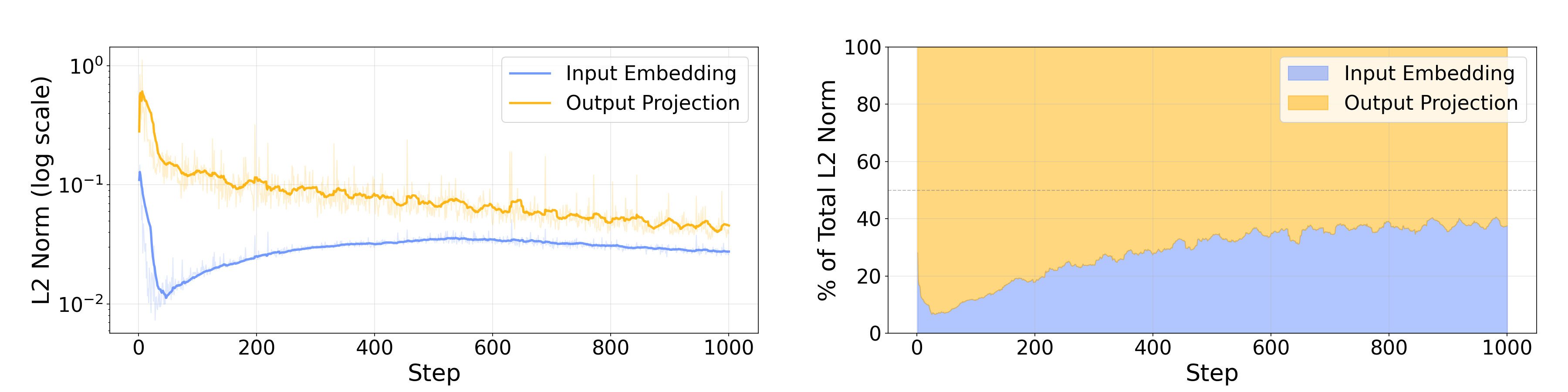}}
    \caption{Gradient flow to the shared embedding matrix in tied OLMo-1B during the first 1000 training steps. Left: L2 norm of gradients from the input embedding (blue) and output projection (orange) pathways on a log scale. Right: relative contribution of each pathway as a percentage of total gradient norm.}
    \label{fig:gradient-provenance}
\end{figure*}

\paragraph{Method} To analyze embedding evolution dynamics, we compute two complementary metrics: 
(1) the mean cosine similarity between each checkpoint and the initial (step 0) embeddings, which measures cumulative drift from initialization, and (2) the mean cosine similarity between each token's embedding vector at consecutive checkpoints, which measures the rate of change at each training step. Figure~\ref{fig:embedding-changes-olmo} shows these metrics for OLMo-1B-0724 (untied) across the first 20000 steps. The plots for Pythia can be found in Figure~\ref{fig:embedding-changes-pythia} (Appendix \ref{app:appendix-matrices-updates}). We sample checkpoints at intervals of 2000 steps for OLMo to ensure meaningful differences between consecutive checkpoints.

\paragraph{Results} Across both models, we observe a consistent pattern: the output embedding matrix undergoes substantially larger updates than the input embedding matrix, especially early in training.

For OLMo-1B-0724 (untied), by step 20K the output matrix cosine similarity with its initialization is approximately 0.56, while for the input matrix it is 0.87 (Figure~\ref{fig:embedding-changes-olmo}). The per-step change rate (bottom panel) reveals that this difference originates in the first few thousand steps. The output matrix shows cosine similarity of 0.96 between consecutive checkpoints, indicating substantial between-checkpoint changes. By contrast, that of the input matrix remains near 1.0, indicating minimal change. Both matrices converge to similar per-step change rates later in training, but the early asymmetry permanently transforms the output matrix. For Pythia-1B, this is even more pronounced (Figure~\ref{fig:embedding-changes-pythia}). The output matrix shifts dramatically from the first checkpoint, with a per-step cosine similarity as low as 0.78 in early training.

We hypothesize that this early-training asymmetry is critical for tied models: \textbf{when input and output embeddings share parameters, the dominant output-layer gradients in the early phases of training shape the shared matrix}, potentially at the expense of early-layer compatibility.

\begin{figure*}[ht]
    \centering
    \includegraphics[width=\textwidth]{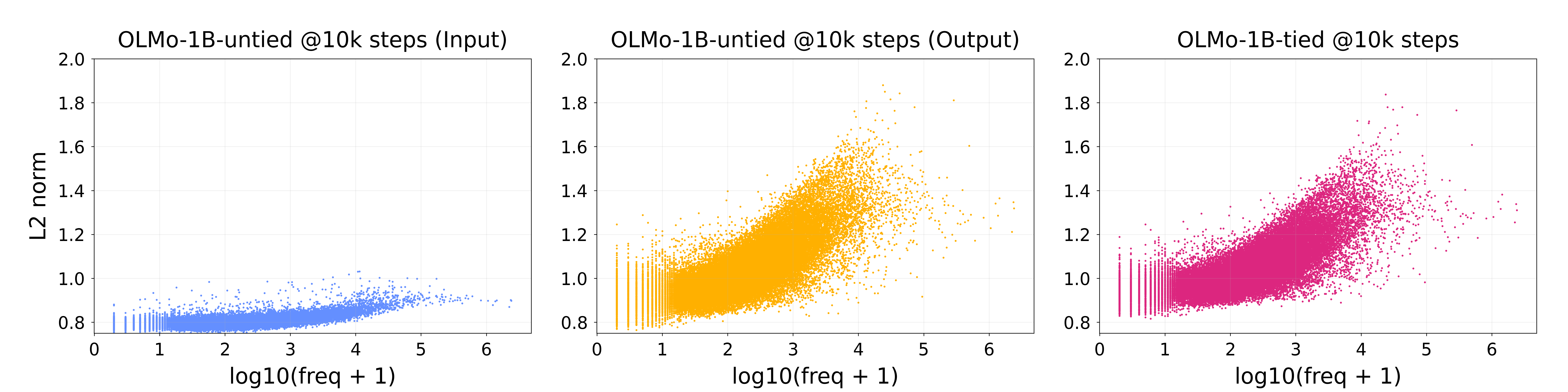}
    \caption{Norm-frequency relationship for OLMo-1B after 10k steps (20B tokens). Left: untied model's input (blue) and output (orange) matrices. Right: tied model's shared matrix (pink).}
    \label{fig:qualitative-comparison}
\end{figure*}

\subsection{Norm-Frequency Relationship} 

We also qualitatively compare the tied and untied embedding vectors using norm-frequency plots, which show how embedding norms vary with token frequency. Here we calculate the norm-distribution of the embedding vectors as a function of the frequency of occurrence of tokens in a corpus.

Figure~\ref{fig:qualitative-comparison} plots the L2 norm of each token's embedding against its log-frequency in a dataset of English text for OLMo-1B after 10k steps (20B tokens). With fixed y-axis scales across all three panels, it is easy to see that the untied input embedding matrix (left) has not moved far from initialization, with norms clustered tightly around 0.8--1.0. In contrast, the untied output (center) and tied (right) matrices have evolved substantially and closely resemble each other, both showing a characteristic pattern where norms increase with token frequency up to approximately $10^4$ occurrences before exhibiting a downward-sloping trend for very high frequency tokens. The visual resemblance between the tied and untied output matrices, at this early stage of training, supports our finding that the tied matrix is shaped predominantly by output-layer gradients from the start. This serves as additional support for our quantitative alignment and model checkpoints analyses above.

\vspace{0.2in}
\subsection{Output Gradients Dominate Training in Weight-Tied Models}
\label{subsec:dominance-results}

The checkpoint analysis in Section~\ref{subsec:dominance-methodology} demonstrates that output embeddings evolve faster than input embeddings when they are separate parameters. But does this gradient asymmetry persist when the matrices are tied? To answer this directly, we measure where gradients originate in weight-tied models. In weight-tied architectures, the same matrix serves both as input embeddings and output projection, with gradients from both pathways accumulating into the shared parameters. To disentangle these contributions without altering training dynamics, we use gradient hooks that observe—but do not modify—the separate gradient flows. We log the L2 norms of gradients through each pathway during early training, quantifying how much learning signal originates from each side.

\paragraph{Method} We train OLMo-1B with tied embeddings from scratch\footnote{\url{https://github.com/allenai/OLMo/tree/main/configs}} using standard OLMo training configuration and record gradient norms at every training step for the first 1000 steps. To reduce noise, we apply rolling average with a window of 20 steps. We do not modify the original training recipe and just track the contributions of gradient flows from output and input embedding paths.

\paragraph{Results} Figure~\ref{fig:gradient-provenance} shows the results. The left panel (log scale) reveals a persistent gap between the two pathways: output-layer gradients (orange) consistently exceed input-layer gradients (blue) throughout the first 1000 training steps. The right panel shows the relative contribution of each pathway (in percent of gradient norm). \textbf{Output-layer gradients account for approximately 70\% of the total signal during early training}, with input-layer contributions making up the remaining 30\%.

This persistent imbalance confirms that in weight-tied architectures, the shared embedding matrix receives the majority of its learning signal from the output layer, particularly during the critical early phase of training when representations are first being shaped. This explains the embedding alignment results from Section \ref{sec:alignment-analysis} and sheds light on the source of the weight tying penalty: \textbf{the shared matrix in weight-tied models is optimized primarily for output prediction from early training stages}, leaving the early transformer layers to work with a less compatible representational space.

\section{Causal Ablation : Selective Gradient Scaling in Tied Models}
\label{sec:gradient-intervention}

Our analysis so far has established that (1) tied embeddings structurally resemble untied output embeddings (Section \ref{subsec:alignment-methodology}), (2) early layers in tied models contribute less effectively to the residual stream (section \ref{subsec:tuned-lens-analysis}), and (3) output-layer gradients dominate the shared embedding matrix during training (\ref{subsec:dominance-results}). In order to establish a causal link showing that gradient flows from output embedding dominate training in tied models, we intervene directly on the shared embedding gradients during training. If output-gradient dominance \emph{causes} the tied matrix to develop output-like structure, then artificially amplifying input-layer gradients should shift the resulting embeddings toward input-like structure. Conversely, if the gradient imbalance is merely a side effect, scaling should have little impact on the final representations. 

We train OLMo-1B with tied embeddings from scratch with the original training configuration, using gradient hooks to multiply input-layer gradients by a scaling factor before they accumulate into the shared embedding matrix. We compare a baseline model with no scaling against one with input gradients scaled by 5$\times$. Since training these models from scratch is expensive, we pause our training for both models after 10000 steps (20B tokens) when the results become evident. We then measure cosine similarity (after alignment) between each tied model's embedding matrix and the input/output matrices from OLMo-1B-0724 (untied) at the same training step. The results are in Table~\ref{tab:gradient-scaling}. We report additional results in Appendix \ref{app:appendix-gradient-scaling}.

\begin{table}[ht]
\centering
\small
\begin{tabular}{lcc}
\toprule
\textbf{Model} & \textbf{vs Untied Input} & \textbf{vs Untied Output} \\
\midrule
Tied (no scaling) & 0.216 & 0.384 \\
Tied (input $\times$5) & 0.222 & 0.369 \\
\bottomrule
\end{tabular}
\caption{Cosine similarity (after alignment) between tied OLMo-1B embedding matrices and the input/output matrices of OLMo-1B-0724 (untied) at step 10K. Higher gradient scaling shifts mean more input-like structure. 
}
\label{tab:gradient-scaling}
\end{table}


\paragraph{Scaling input gradients shifts embedding structure predictably towards untied input embeddings and away from untied output embeddings.} With 5$\times$ input gradient scaling, the resulting embedding matrix becomes more aligned with the untied input matrix (0.216 $\rightarrow$ 0.222) but is \textit{less} aligned with the untied output matrix (0.384 $\rightarrow$ 0.369). Additional scaling factors and earlier checkpoints show the same trend (Appendix~\ref{app:appendix-gradient-scaling}).

This trade-off is inherent to weight tying. Shifting the shared matrix toward one role necessarily moves it away from the other, and the gradient balance causally determines which role dominates. Under this simple constant-scaling intervention, there is no ``free'' equilibrium, improving input compatibility comes at the cost of output compatibility. Downstream evaluation shows no consistent performance gains from this structural shift (Appendix~\ref{app:appendix-gradient-scaling}), which is in line with the zero-sum nature of this trade-off. Whether more sophisticated interventions can navigate this trade-off remains an open question.

\section{Discussion}
\label{sec:discussion}

\subsection{When to Tie vs. Untie}
\label{subsec:when-to-tie}

The decision to tie or untie embeddings ultimately reduces to a trade-off between parameter efficiency and representational capacity. Above, we highlighted that in smaller models, embedding parameters may constitute 70\% of total parameters (Pythia 70M; see Table~\ref{tab:embedding-params}). At this scale, these parameter savings are substantial and justify the representational cost we have documented. At $>$1B parameters, the efficiency advantage largely disappears while the representational disadvantage remains. This trade-off is reflected in contemporary models. The Qwen3 developers explicitly cite this as a consideration: smaller models (0.6B, 1.7B, and 4B parameters) use tied embeddings, while larger models (8B and above) untie them. Here, we  provide a mechanistic justification for this threshold: at scale, the cost of adding a separate output matrix is negligible, but the benefit of allowing the input embeddings to evolve independently is present.

\subsection{Gradient Rebalancing as a Training Intervention}
\label{subsec:gradient-rebalancing}

Our causal intervention experiments (Section~\ref{sec:gradient-intervention}) establish that gradient imbalance causally determines the structure of the tied embedding matrix. We also evaluated whether rebalancing gradients improves downstream performance and find that it does not: the scaled model shows no consistent gains over the baseline. This is in line with the trade-off we document, as shifting the shared matrix toward input-like structure necessarily reduces its effectiveness as an output projection.

Future work could explore whether more targeted interventions, beyond simple constant scaling, can navigate this trade-off. If such strategies exist, they could offer a practical middle ground: tied embeddings with intervention-corrected training dynamics that mitigate the representational cost we have documented while retaining the parameter efficiency of weight tying. \textit{This can have important implications for smaller language models where embeddings form a large portion of the parameters.}

\section{Conclusion}
\label{sec:conclusion}

We have presented evidence that weight tying in language models incurs a hidden cost. By analyzing training dynamics through gradient flow, we show that the shared embedding matrix in tied models is shaped predominantly by output-layer gradients during early training, leaving input representations optimized for output prediction rather than for compatibility with early-layer computations.

This gradient imbalance manifests as a representational mismatch. We show that tied embedding matrices are structurally closer to untied output matrices than to untied input matrices, confirming that output-layer optimization dominates the shared representation. Furthermore, in our tuned lens analysis across three model families, tied models exhibit systematically higher KL divergence in their early layers compared to untied counterparts, indicating that the first layers contribute less effectively to the residual stream, a pattern consistent with input embeddings that have been trained suboptimally compared to the untied models. Untying allows the input matrix to evolve more slowly than the output matrix, free from the dominant output-layer gradients that would otherwise shape it.

Our findings provide a mechanistic explanation for the empirical observation that untying embeddings improves performance at scale. At billion-scale and beyond, the parameter savings from weight tying become negligible while the representational cost remains. For practitioners designing large language models, we recommend untying embeddings when parameter budget permits, as the representational benefits outweigh the modest increase in model size. For smaller models where parameter savings matter most, our work motivates training interventions that could preserve the efficiency benefits of weight tying while reducing its representational cost.

\section*{Limitations}
\label{sec:limitations}

Our analysis relies on several assumptions that merit discussion. While the tuned lens reveals that tied models exhibit systematically higher KL divergence in early layers compared to untied models, we cannot be certain this difference reflects a meaningful representational cost. The elevated divergence may be compensated for in later layers.

For each setting, we test only a single training run, as training runs at the billion-parameter scale are very costly, therefore we were not able to conduct statistical tests across multiple training runs, random seeds, etc. 

Our experiments are also constrained to specific model families (OLMo, Pythia, and Qwen3), chosen for their availability of tied/untied variants or intermediate checkpoints. While these models span different training configurations and scales, our findings may not generalize to architectures with substantially different designs, such as mixture-of-experts models or non-autoregressive language models.


\bibliography{custom}

@article{duderstadt2023comparing,
    title={Comparing Foundation Models using Data Kernels},
    author={Duderstadt, Brandon and Helm, Hayden S. and Priebe, Carey E.},
    journal={arXiv preprint arXiv:2305.05126},
    year={2023},
    url={https://arxiv.org/abs/2305.05126}
}

@article{chung2020rethinking,
    title={Rethinking embedding coupling in pre-trained language models},
    author={Chung, Hyung Won and F{\'e}vry, Thibault and Tsai, Henry and Johnson, Melvin and Ruder, Sebastian},
    journal={arXiv preprint arXiv:2010.12821},
    year={2020},
    url={https://arxiv.org/abs/2010.12821}
}

@article{hu2020xtreme,
    title={{XTREME}: A Massively Multilingual Multi-task Benchmark for Evaluating Cross-lingual Generalization},
    author={Hu, Junjie and Ruder, Sebastian and Siddhant, Aditya and Neubig, Graham and Firat, Orhan and Johnson, Melvin},
    journal={arXiv preprint arXiv:2003.11080},
    year={2020},
    url={https://arxiv.org/abs/2003.11080}
}

@article{groeneveld2024olmo,
    title={OLMo: Accelerating the Science of Language Models},
    author={Groeneveld, Dirk and Beltagy, Iz and Walsh, Pete and Bhagia, Akshita and Kinney, Rodney and Tafjord, Oyvind and Jha, Ananya Harsh and Ivison, Hamish and Magnusson, Ian and Wang, Yizhong and others},
    journal={arXiv preprint arXiv:2402.00838},
    year={2024},
    url={https://arxiv.org/abs/2402.00838}
}

@article{olmo2furious,
    title={OLMo 2: The best fully open language model to date},
    author={{Team OLMo} and Walsh, Pete and Soldaini, Luca and Groeneveld, Dirk and Lo, Kyle and Arora, Shane and Bhagia, Akshita and Gu, Yuling and Huang, Shengyi and Jordan, Matt and Lambert, Nathan and Schwenk, Dustin and Tafjord, Oyvind and others},
    journal={arXiv preprint arXiv:2501.00656},
    year={2024},
    url={https://arxiv.org/abs/2501.00656}
}

@article{olmo3,
    title={Olmo 3},
    author={{Team Olmo} and Ettinger, Allyson and Bertsch, Amanda and Kuehl, Bailey and Graham, David and Heineman, David and Groeneveld, Dirk and Brahman, Faeze and Timbers, Finbarr and Ivison, Hamish and others},
    journal={arXiv preprint arXiv:2512.13961},
    year={2025},
    url={https://arxiv.org/abs/2512.13961}
}

@inproceedings{machina2024anisotropy,
    title={Anisotropy is Not Inherent to Transformers},
    author={Machina, Anemily and Mercer, Robert},
    booktitle={Proceedings of the 2024 Conference of the North American Chapter of the Association for Computational Linguistics: Human Language Technologies (Volume 1: Long Papers)},
    year={2024},
    publisher={Association for Computational Linguistics},
    url={https://aclanthology.org/2024.naacl-long.274/}
}

@inproceedings{rajaee2022isotropy,
    title={An Isotropy Analysis in the Multilingual {BERT} Embedding Space},
    author={Rajaee, Sara and Pilehvar, Mohammad Taher},
    booktitle={Findings of the Association for Computational Linguistics: ACL 2022},
    year={2022},
    publisher={Association for Computational Linguistics},
    url={https://aclanthology.org/2022.findings-acl.103/}
}

@article{biderman2023pythia,
    title={Pythia: A Suite for Analyzing Large Language Models Across Training and Scaling},
    author={Biderman, Stella and Schoelkopf, Hailey and Anthony, Quentin and Bradley, Herbie and O'Brien, Kyle and Hallahan, Eric and Khan, Mohammad Aflah and Purohit, Shivanshu and Prashanth, USVSN Sai and Raff, Edward and others},
    journal={arXiv preprint arXiv:2304.01373},
    year={2023},
    url={https://arxiv.org/abs/2304.01373}
}

@article{black2022gptneox,
    title={GPT-NeoX-20B: An Open-Source Autoregressive Language Model},
    author={Black, Sid and Biderman, Stella and Hallahan, Eric and Anthony, Quentin and Gao, Leo and Golber, Laurence and He, Horace and Leahy, Connor and McDonell, Kyle and Phang, Jason and others},
    journal={arXiv preprint arXiv:2204.06745},
    year={2022},
    url={https://arxiv.org/abs/2204.06745}
}

@misc{qwen3technicalreport,
    title={Qwen3 Technical Report},
    author={{Qwen Team}},
    year={2025},
    eprint={2505.09388},
    archivePrefix={arXiv},
    primaryClass={cs.CL},
    url={https://arxiv.org/abs/2505.09388}
}

@article{belrose2023tunedlens,
    title={Eliciting Latent Predictions from Transformers with the Tuned Lens},
    author={Belrose, Nora and Furman, Zach and Smith, Logan and Halawi, Danny and Ostrovsky, Igor and McKinney, Lev and Biderman, Stella and Steinhardt, Jacob},
    journal={arXiv preprint arXiv:2303.08112},
    year={2023},
    url={https://arxiv.org/abs/2303.08112}
}

@article{liu2019roberta,
    title={{RoBERTa}: A Robustly Optimized {BERT} Pretraining Approach},
    author={Liu, Yinhan and Ott, Myle and Goyal, Naman and Du, Jingfei and Joshi, Mandar and Chen, Danqi and Levy, Omer and Lewis, Mike and Zettlemoyer, Luke and Stoyanov, Veselin},
    journal={arXiv preprint arXiv:1907.11692},
    year={2019},
    url={https://arxiv.org/abs/1907.11692}
}

@article{team2024gemma,
    title={Gemma: Open Models Based on {Gemini} Research and Technology},
    author={{Gemma Team} and Mesnard, Thomas and Hardin, Cassidy and Dadashi, Robert and Bhupatiraju, Surya and Pathak, Shreya and Sifre, Laurent and Rivi{\`e}re, Morgane and Kale, Mihir Sanjay and Love, Juliette and others},
    journal={arXiv preprint arXiv:2403.08295},
    year={2024},
    url={https://arxiv.org/abs/2403.08295}
}

@article{grattafiori2024llama3,
    title={The {Llama 3} Herd of Models},
    author={Grattafiori, Aaron and Dubey, Abhimanyu and Jauhri, Abhinav and Pandey, Abhinav and Kadian, Abhishek and Al-Dahle, Ahmad and Letman, Aiesha and Mathur, Akhil and Schelten, Alan and Yang, Amy and others},
    journal={arXiv preprint arXiv:2407.21783},
    year={2024},
    url={https://arxiv.org/abs/2407.21783}
}

@article{deepseekai2024deepseekv3,
    title={{DeepSeek-V3} Technical Report},
    author={{DeepSeek-AI}},
    journal={arXiv preprint arXiv:2412.19437},
    year={2024},
    url={https://arxiv.org/abs/2412.19437}
}

@misc{nostalgebraist2020logitlens,
    title={interpreting {GPT}: the logit lens},
    author={nostalgebraist},
    year={2020},
    howpublished={LessWrong},
    url={https://www.lesswrong.com/posts/AcKRB8wDpdaN6v6ru/interpreting-gpt-the-logit-lens}
}

@article{mikolov2013exploiting,
    title={Exploiting similarities among languages for machine translation},
    author={Mikolov, Tomas and Le, Quoc V and Sutskever, Ilya},
    journal={arXiv preprint arXiv:1309.4168},
    year={2013},
    url={https://arxiv.org/abs/1309.4168}
}

@inproceedings{xing2015normalized,
    title={Normalized Word Embedding and Orthogonal Transform for Bilingual Word Translation},
    author={Xing, Chao and Wang, Dong and Liu, Chao and Lin, Yiye},
    booktitle={Proceedings of the 2015 Conference of the North American Chapter of the Association for Computational Linguistics: Human Language Technologies},
    pages={1006--1011},
    year={2015},
    publisher={Association for Computational Linguistics},
    url={https://aclanthology.org/N15-1104/}
}

@inproceedings{artetxe2016learning,
    title={Learning principled bilingual mappings of word embeddings while preserving monolingual invariance},
    author={Artetxe, Mikel and Labaka, Gorka and Agirre, Eneko},
    booktitle={Proceedings of the 2016 Conference on Empirical Methods in Natural Language Processing},
    pages={2289--2294},
    year={2016},
    publisher={Association for Computational Linguistics},
    url={https://aclanthology.org/D16-1250/}
}

@inproceedings{conneau2020emerging,
    title={Emerging Cross-lingual Structure in Pretrained Language Models},
    author={Conneau, Alexis and Wu, Shijie and Li, Haoran and Zettlemoyer, Luke and Stoyanov, Veselin},
    booktitle={Proceedings of the 58th Annual Meeting of the Association for Computational Linguistics},
    pages={6022--6034},
    year={2020},
    publisher={Association for Computational Linguistics},
    url={https://aclanthology.org/2020.acl-main.536/}
}

@inproceedings{press2017using,
    title={Using the Output Embedding to Improve Language Models},
    author={Press, Ofir and Wolf, Lior},
    booktitle={Proceedings of the 15th Conference of the European Chapter of the Association for Computational Linguistics: Volume 2, Short Papers},
    pages={157--163},
    year={2017},
    organization={Association for Computational Linguistics},
    url={https://arxiv.org/abs/1608.05859}
}

@inproceedings{inan2017tying,
    title={Tying Word Vectors and Word Classifiers: A Loss Framework for Language Modeling},
    author={Inan, Hakan and Khosravi, Khashayar and Socher, Richard},
    booktitle={Proceedings of ICLR},
    year={2017}
}

@inproceedings{ethayarajh2019contextual,
    title={How Contextual are Contextualized Word Representations? Comparing the Geometry of {BERT}, {ELM}o, and {GPT}-2 Embeddings},
    author={Ethayarajh, Kawin},
    booktitle={Proceedings of EMNLP-IJCNLP},
    pages={55--65},
    year={2019}
}

@techreport{radford2019language,
    title={Language Models are Unsupervised Multitask Learners},
    author={Radford, Alec and Wu, Jeffrey and Child, Rewon and Luan, David and Amodei, Dario and Sutskever, Ilya},
    institution={OpenAI},
    year={2019}
}

@inproceedings{devlin2019bert,
    title={{BERT}: Pre-training of Deep Bidirectional Transformers for Language Understanding},
    author={Devlin, Jacob and Chang, Ming-Wei and Lee, Kenton and Toutanova, Kristina},
    booktitle={Proceedings of NAACL-HLT},
    pages={4171--4186},
    year={2019}
}

@article{schonemann1966,
    title={A Generalized Solution of the Orthogonal Procrustes Problem},
    author={Sch{\"o}nemann, Peter H},
    journal={Psychometrika},
    volume={31},
    number={1},
    pages={1--10},
    year={1966}
}

@article{soldaini2024dolma,
    title={Dolma: An Open Corpus of Three Trillion Tokens for Language Model Pretraining Research},
    author={Soldaini, Luca and Kinney, Rodney and Bhagia, Akshita and Schwenk, Dustin and Atkinson, David and Authur, Russell and Bogin, Ben and Chandu, Khyathi and Dumas, Jennifer and Elazar, Yanai and others},
    journal={arXiv preprint arXiv:2402.00159},
    year={2024}
}

@inproceedings{gao2019representation,
    title={Representation Degeneration Problem in Training Natural Language Generation Models},
    author={Gao, Jun and He, Di and Tan, Xu and Qin, Tao and Wang, Liwei and Liu, Tie-Yan},
    booktitle={Proceedings of ICLR},
    year={2019},
    url={https://arxiv.org/abs/1907.12009}
}

@inproceedings{wang2019adversarial,
    title={Improving Neural Language Modeling via Adversarial Training},
    author={Wang, Dilin and Gong, Chengyue and Liu, Qiang},
    booktitle={Proceedings of the 36th International Conference on Machine Learning},
    pages={6555--6565},
    year={2019},
    volume={97},
    series={Proceedings of Machine Learning Research},
    publisher={PMLR},
    url={https://proceedings.mlr.press/v97/wang19f.html}
}

@inproceedings{wang2020spectrum,
    title={Improving Neural Language Generation with Spectrum Control},
    author={Wang, Lingxiao and Huang, Jing and Huang, Kevin and Hu, Ziniu and Wang, Guangtao and Gu, Quanquan},
    booktitle={International Conference on Learning Representations},
    year={2020},
    url={https://openreview.net/forum?id=ByxY8CNtvr}
}

@article{stollenwerk2025coupled,
    title={Better Embeddings with Coupled Adam},
    author={Stollenwerk, Felix and Stollenwerk, Tobias},
    journal={arXiv preprint arXiv:2502.08441},
    year={2025},
    url={https://arxiv.org/abs/2502.08441}
}

@inproceedings{bis2021toomuch,
    title={Too Much in Common: Shifting of Embeddings in Transformer Language Models and its Implications},
    author={Bi{\'s}, Daniel and Podkorytov, Maksim and Liu, Xiuwen},
    booktitle={Proceedings of the 2021 Conference of the North American Chapter of the Association for Computational Linguistics: Human Language Technologies},
    pages={5117--5130},
    year={2021},
    publisher={Association for Computational Linguistics},
    url={https://aclanthology.org/2021.naacl-main.403/}
}

@article{elman1990finding,
    title={Finding Structure in Time},
    author={Elman, Jeffrey L},
    journal={Cognitive Science},
    volume={14},
    number={2},
    pages={179--211},
    year={1990},
    publisher={Wiley}
}

@inproceedings{vaswani2017attention,
    title={Attention Is All You Need},
    author={Vaswani, Ashish and Shazeer, Noam and Parmar, Niki and Uszkoreit, Jakob and Jones, Llion and Gomez, Aidan N and Kaiser, {\L}ukasz and Polosukhin, Illia},
    booktitle={Advances in Neural Information Processing Systems},
    volume={30},
    year={2017}
}

@article{gao2020pile,
    title={The Pile: An 800GB Dataset of Diverse Text for Language Modeling},
    author={Gao, Leo and Biderman, Stella and Black, Sid and Golding, Laurence and Hoppe, Travis and Foster, Charles and Phang, Jason and He, Horace and Thite, Anish and Nabeshima, Noa and Presser, Shawn and Leahy, Connor},
    journal={arXiv preprint arXiv:2101.00027},
    year={2020}
}

@inproceedings{bertolotti2024tying,
    title={By Tying Embeddings You Are Assuming the Distributional Hypothesis},
    author={Bertolotti, Francesco and Cazzola, Walter},
    booktitle={Proceedings of the 41st International Conference on Machine Learning},
    pages={3584--3610},
    year={2024},
    volume={235},
    series={Proceedings of Machine Learning Research},
    publisher={PMLR},
    url={https://proceedings.mlr.press/v235/bertolotti24a.html}
}

@inproceedings{merity2017pointer,
    title={Pointer Sentinel Mixture Models},
    author={Merity, Stephen and Xiong, Caiming and Bradbury, James and Socher, Richard},
    booktitle={International Conference on Learning Representations},
    year={2017},
    url={https://arxiv.org/abs/1609.07843}
}

@inproceedings{bisk2020piqa,
    title={{PIQA}: Reasoning about Physical Commonsense in Natural Language},
    author={Bisk, Yonatan and Zellers, Rowan and Le Bras, Ronan and Gao, Jianfeng and Choi, Yejin},
    booktitle={Proceedings of the AAAI Conference on Artificial Intelligence},
    volume={34},
    number={5},
    pages={7432--7439},
    year={2020}
}

@inproceedings{zellers2019hellaswag,
    title={{HellaSwag}: Can a Machine Really Finish Your Sentence?},
    author={Zellers, Rowan and Holtzman, Ari and Bisk, Yonatan and Farhadi, Ali and Choi, Yejin},
    booktitle={Proceedings of the 57th Annual Meeting of the Association for Computational Linguistics},
    pages={4791--4800},
    year={2019},
    url={https://aclanthology.org/P19-1472/}
}

@inproceedings{sakaguchi2020winogrande,
    title={{WinoGrande}: An Adversarial Winograd Schema Challenge at Scale},
    author={Sakaguchi, Keisuke and Le Bras, Ronan and Bhagavatula, Chandra and Choi, Yejin},
    booktitle={Proceedings of the AAAI Conference on Artificial Intelligence},
    volume={34},
    number={5},
    pages={8732--8740},
    year={2020}
}

@article{clark2018arc,
    title={Think you have Solved Question Answering? {Try ARC}, the {AI2} Reasoning Challenge},
    author={Clark, Peter and Cowhey, Isaac and Etzioni, Oren and Khot, Tushar and Sabharwal, Ashish and Schoenick, Carissa and Tafjord, Oyvind},
    journal={arXiv preprint arXiv:1803.05457},
    year={2018},
    url={https://arxiv.org/abs/1803.05457}
}

@inproceedings{clark2019boolq,
    title={{BoolQ}: Exploring the Surprising Difficulty of Natural Yes/No Questions},
    author={Clark, Christopher and Lee, Kenton and Chang, Ming-Wei and Kwiatkowski, Tom and Collins, Michael and Toutanova, Kristina},
    booktitle={Proceedings of the 2019 Conference of the North American Chapter of the Association for Computational Linguistics: Human Language Technologies},
    pages={2924--2936},
    year={2019},
    url={https://aclanthology.org/N19-1300/}
}

@inproceedings{mihaylov2018openbookqa,
    title={Can a Suit of Armor Conduct Electricity? {A} New Dataset for Open Book Question Answering},
    author={Mihaylov, Todor and Clark, Peter and Khot, Tushar and Sabharwal, Ashish},
    booktitle={Proceedings of the 2018 Conference on Empirical Methods in Natural Language Processing},
    pages={2381--2391},
    year={2018},
    url={https://aclanthology.org/D18-1260/}
}

@misc{eval-harness,
    author={Gao, Leo and Tow, Jonathan and Abbasi, Baber and Biderman, Stella and Black, Sid and DiPofi, Anthony and Foster, Charles and Golding, Laurence and Hsu, Jeffrey and Le Noac'h, Alain and Li, Haonan and McDonell, Kyle and Muennighoff, Niklas and Ociepa, Chris and Phang, Jason and Reynolds, Laria and Schoelkopf, Hailey and Skowron, Aviya and Sutawika, Lintang and Tang, Eric and Thite, Anish and Wang, Ben and Wang, Kevin and Zou, Andy},
    title={A framework for few-shot language model evaluation},
    month={12},
    year={2023},
    publisher={Zenodo},
    version={v0.4.0},
    doi={10.5281/zenodo.10256836},
    url={https://zenodo.org/records/10256836}
}

@article{warstadt2020blimp,
    title={{BLiMP}: The Benchmark of Linguistic Minimal Pairs for {E}nglish},
    author={Warstadt, Alex and Parrish, Alicia and Liu, Haokun and Mohananey, Anhad and Peng, Wei and Wang, Sheng-Fu and Bowman, Samuel R.},
    journal={Transactions of the Association for Computational Linguistics},
    volume={8},
    pages={377--392},
    year={2020},
    url={https://aclanthology.org/2020.tacl-1.25/}
}

\appendix
\appendix
\section*{Appendix}
\addcontentsline{toc}{section}{Appendix}
\section{Parameter Counts} \label{app:param_counts}

Table~\ref{tab:embedding-params} shows the proportion of total parameters contained in the embedding matrices for the Pythia model family \citep{biderman2023pythia}. This family of models does not use weight tying. For the 70M parameter model, the embedding matrices account for over 50\% of all parameters; even at 2.8B scale, they still represent approximately 7\% of the model, so tying can still yield meaningful savings.

\begin{table}[t]
\centering
\small
\setlength{\tabcolsep}{5pt}
\begin{tabular}{lrrr}
\toprule
\textbf{Model} & \textbf{Total (Untied)} & \textbf{Total (Tied)} & \textbf{Saved} \\
\midrule
Pythia-70M  & 70.4M   & 44.7M   & 25.8M \\
Pythia-160M & 162.3M  & 123.7M  & 38.6M \\
Pythia-410M & 405.3M  & 353.8M  & 51.5M \\
Pythia-1B   & 1011.8M & 908.8M  & 103.0M \\
Pythia-2.8B & 2775.2M & 2646.4M & 128.8M \\
\bottomrule
\end{tabular}
\caption{Total parameter counts with and without weight tying for Pythia models. For an untied model, tying removes one embedding matrix, saving $Vd$ parameters (half of the untied embedding total $2Vd$). Totals are computed by summing checkpoint parameter counts.}
\label{tab:pythia-tied-untied-totals}
\end{table}

Table~\ref{tab:pythia-tied-untied-totals} reports total parameter counts with and without weight tying, computed by subtracting $Vd$ parameters from the untied checkpoints.

Table~\ref{tab:embedding-params} shows that embedding matrices constitute a substantial fraction of total parameters at smaller scales, but this proportion decreases as models grow. At 70M parameters, embeddings account for 73.4\% of the model, making weight tying highly impactful for parameter efficiency. By 2.8B parameters, this drops to 9.2\%.

\begin{table}[t]
\centering
\small
\begin{tabular}{lrrr}
\toprule
\textbf{Model} & \textbf{Hidden} & \textbf{Embed Params} & \textbf{\% of Total} \\
\midrule
Pythia-70M & 512 & 51.4M & 73.4\% \\
Pythia-160M & 768 & 77.2M & 48.3\% \\
Pythia-410M & 1024 & 103.0M & 25.1\% \\
Pythia-1B & 2048 & 206.0M & 20.6\% \\
Pythia-2.8B & 2560 & 257.4M & 9.2\% \\
\bottomrule
\end{tabular}
\caption{Embedding parameters for the Pythia suite (untied). Embed Params shows the combined input and output matrices ($2 \times V \times d$ where $V = 50{,}257$). Weight tying would halve these values, saving over 35\% of parameters at small scale but less than 5\% at billion-scale.}
\label{tab:embedding-params}
\end{table}

\section{KNN Overlap and Spectral Distance Analysis}
\label{app:appendix-knn}

As complementary metrics to cosine similarity, we measure (1)~\textbf{KNN@10 overlap}: the fraction of shared nearest neighbors between token embeddings in two matrices (higher = more similar), and (2)~\textbf{spectral distance}: the operator-norm distance between the jointly embedded k-NN graphs via the omnibus adjacency spectral embedding of \citet{duderstadt2023comparing} (lower = more similar).

\begin{table}[ht]
\centering
\small
\setlength{\tabcolsep}{3pt}
\begin{tabular}{lcc}
\toprule
\textbf{Comparison} & \textbf{KNN@10} ($\uparrow$) & \textbf{Spectral Dist} ($\downarrow$) \\
\midrule
\multicolumn{3}{l}{\textit{OLMo-1B (tied) vs OLMo-1B-0724 (untied)}} \\
Input (U) $\leftrightarrow$ Output (U) & 0.455          & 1.349          \\
Output (U) $\leftrightarrow$ Tied      & \textbf{0.733} & \textbf{0.588} \\
Input (U) $\leftrightarrow$ Tied       & 0.496          & 1.373          \\
\midrule
\multicolumn{3}{l}{\textit{Qwen3-4B (tied) vs Qwen3-8B (untied)}} \\
Input (U) $\leftrightarrow$ Output (U) & 0.366          & 1.421          \\
Output (U) $\leftrightarrow$ Tied      & \textbf{0.710} & \textbf{0.809} \\
Input (U) $\leftrightarrow$ Tied       & 0.366          & 1.441          \\
\midrule
\multicolumn{3}{l}{\textit{GPT-Neo-2.7B (tied) vs Pythia-2.8B (untied)}} \\
Input (U) $\leftrightarrow$ Output (U) & \textbf{0.611} & \textbf{0.949} \\
Output (U) $\leftrightarrow$ Tied      & 0.408          & 1.035          \\
Input (U) $\leftrightarrow$ Tied       & 0.372          & 1.462          \\
\bottomrule
\end{tabular}
\caption{KNN@10 overlap (higher = more similar) and spectral distance (lower = more similar) between embedding spaces. Bold indicates the most similar pair within each family. GPT-Neo/Pythia uses aligned vocabulary (36,938 tokens).}
\label{tab:knn-alignment}
\end{table}

Both metrics tell a consistent story. For OLMo-1B, the tied matrix shares 73.3\% of its nearest neighbors with the untied output versus only 49.6\% with the untied input; spectral distances are 0.588 versus 1.373. Qwen3 shows an even starker contrast: 71.0\% overlap with output versus 36.6\% with input; spectral distances are 0.809 versus 1.441. The pattern holds for GPT-Neo/Pythia (KNN@10: 0.408 vs.\ 0.372; spectral distance: 1.035 vs.\ 1.462), though the effect is smaller, likely because of the bigger differences between the two models.

\section{Additional Results --- Tuned Lens Analysis}
\label{app:appendix-tuned-lens}

Section~\ref{subsec:tuned-lens-analysis} presented tuned lens results for OLMo-1B (tied vs untied). Here we extend this analysis to Pythia-2.8B vs GPT-Neo-2.7B and Qwen3-4B vs Qwen3-8B. The methodology remains identical: we train per-layer affine translators that map hidden states at each layer to the final layer's representation space, minimizing KL divergence. 

\begin{figure}[ht]
    \centering
    \includegraphics[width=\columnwidth]{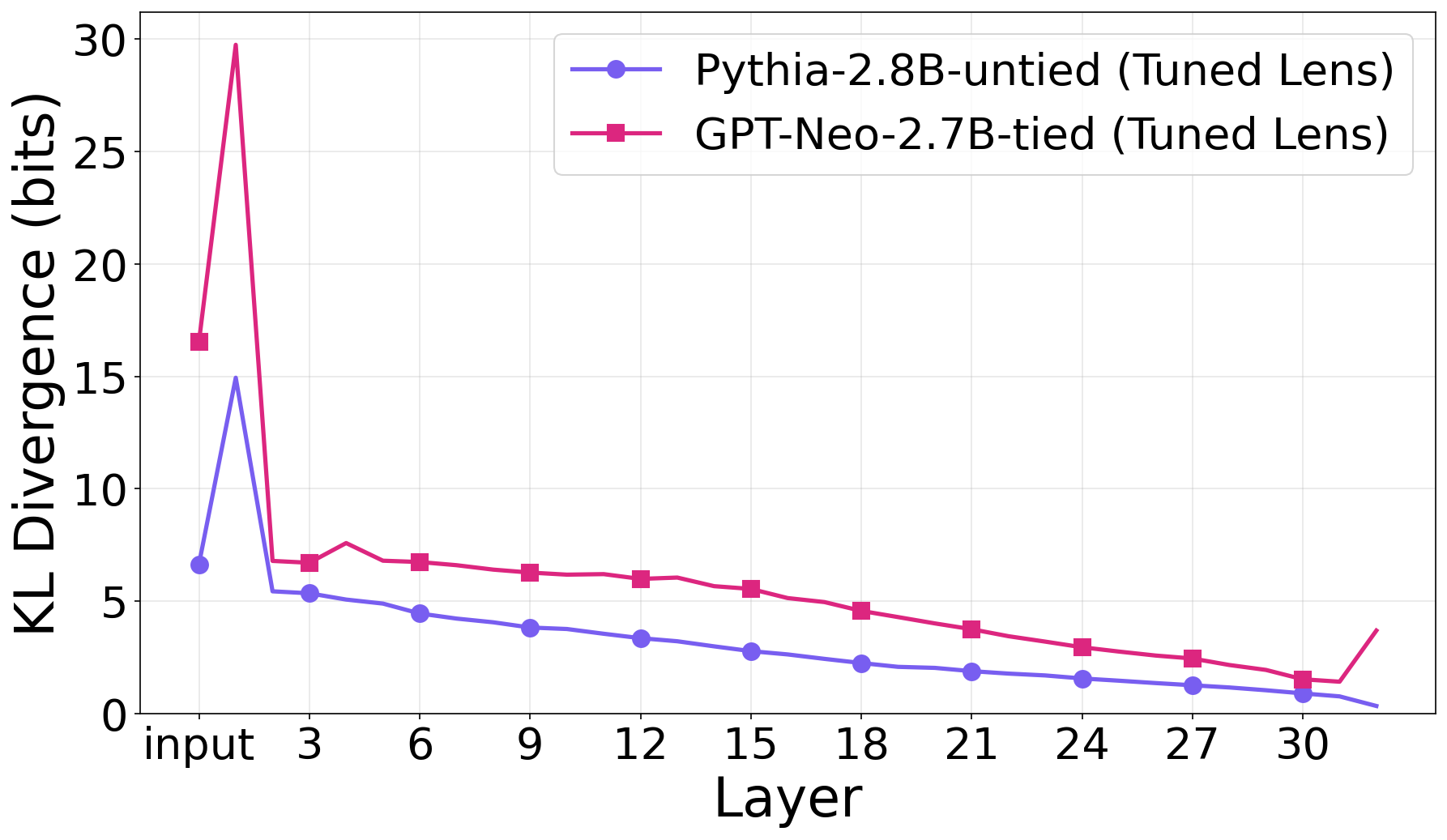}
    \caption{Tuned lens KL divergence across layers for Pythia-2.8B (untied) vs GPT-Neo-2.7B (tied). Both models have 32 layers. GPT-Neo-2.7B shows elevated KL divergence in the early layers.}
    \label{fig:pythia-vs-gptneo}
\end{figure}

Figure~\ref{fig:pythia-vs-gptneo} shows results for Pythia-2.8B (untied) and GPT-Neo-2.7B (tied). Both models have 32 layers. GPT-Neo-2.7B (tied) starts at approximately 17 bits at the input layer compared to 7 bits for Pythia-2.8B (untied), a gap of 10 bits. This gap subsequently increases to 15 bits and then narrows, with the curves starting to converge around layer 10. Both models show a spike in KL divergence early on followed by a smooth, monotonic reduction until convergence. Pythia-2.8B (untied) also exhibits a slight increase from the input layer to layer 1, though much smaller than the tied model's spike. The elevated early-layer KL divergence in GPT-Neo-2.7B suggests that early layer activations in 
tied models are harder to align with the output space and the first layer is contributing less effectively to the residual stream.

\begin{figure}[ht]
    \centering
    \includegraphics[width=\columnwidth]{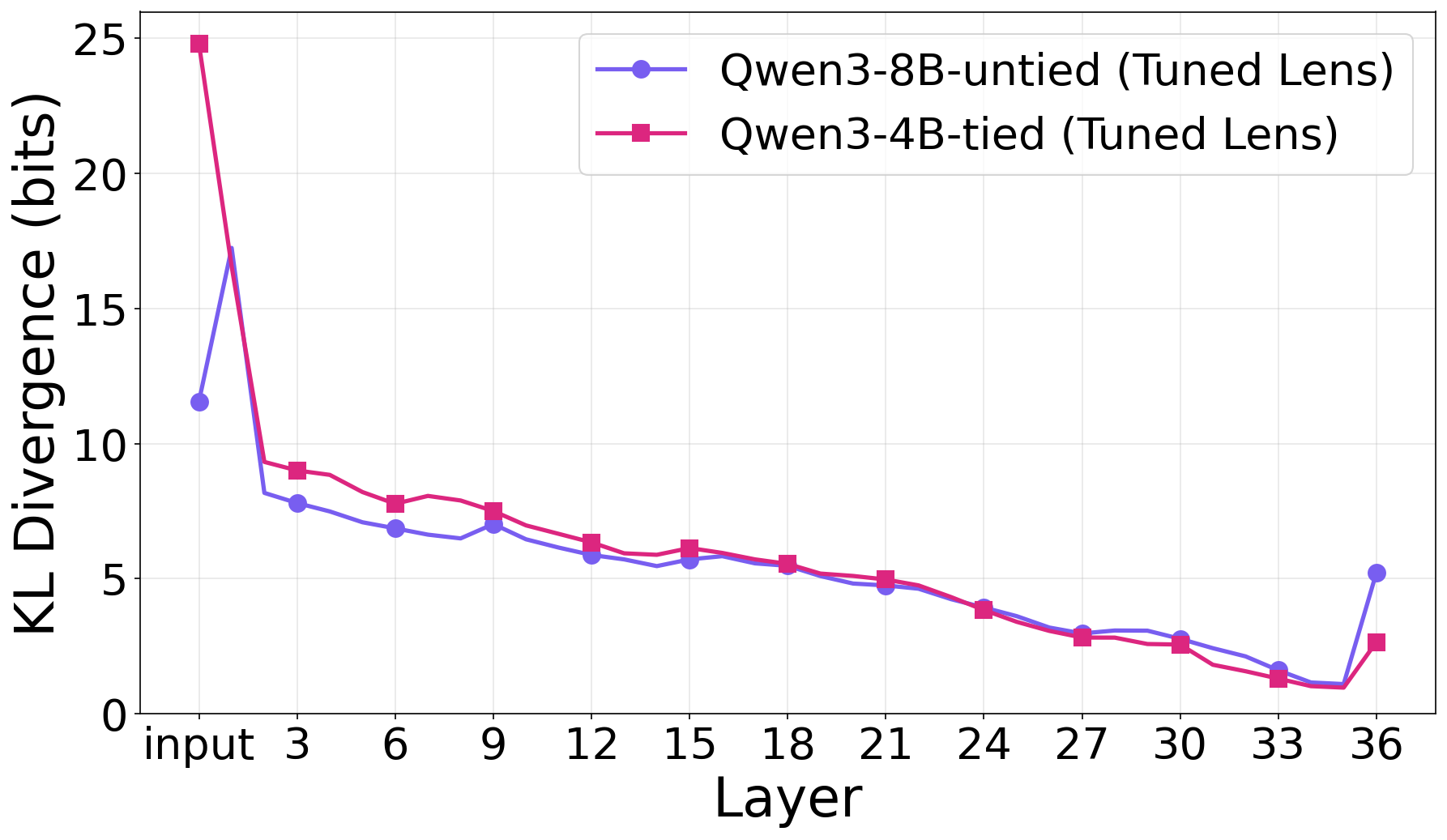}
    \caption{Tuned lens KL divergence for Qwen3-4B (tied) vs Qwen3-8B (untied). Qwen3-4B (tied) displays significantly higher early-layer KL divergence before starting to converge with Qwen3-8B (untied).}
    \label{fig:qwen3-comparison}
\end{figure}

Figure~\ref{fig:qwen3-comparison} presents results for Qwen3-4B (tied, 36 layers) and Qwen3-8B (untied, 36 layers). We again observe tied models exhibiting higher early-layer KL divergence. Qwen3-4B (tied) begins at approximately 25 bits, while Qwen3-8B (untied) starts at 12 bits, a 13-bit gap. Qwen3-8B (untied) also shows a slight increase from the input layer to layer 1, though the gap relative to the tied model remains large. This gap narrows quickly, with the tied model maintaining elevated divergence through layer 10, where the curves start to converge. In the final layers, both models show increasing KL divergence.

The consistency of elevated early-layer KL divergence across three independent model family comparisons in OLMo (same family, different training runs), Pythia/GPT-Neo (similar architecture, same dataset), and Qwen3 (same family, different sizes), provides evidence that weight tying creates an early-layer representational penalty. This pattern holds across different model scales (from 1B to 8B parameters), different training datasets, and different architectural details, even though the exact depth profile beyond the earliest layers varies across architectures.


\section{Additional Results --- Untied Models Matrices Updates}
\label{app:appendix-matrices-updates}

Section~\ref{subsec:dominance-methodology} presented embedding evolution dynamics for OLMo-1B-0724 (untied). Here we extend this to OLMo-7B and Pythia-1B. For each model, we compute cumulative drift (cosine similarity to initial embeddings) and per-step change rate (cosine similarity between consecutive checkpoints). OLMo checkpoints are available every 1000 steps; Pythia has 154 checkpoints spanning the full training run.

\begin{figure}[htbp]
    \centering
    \includegraphics[width=\columnwidth]{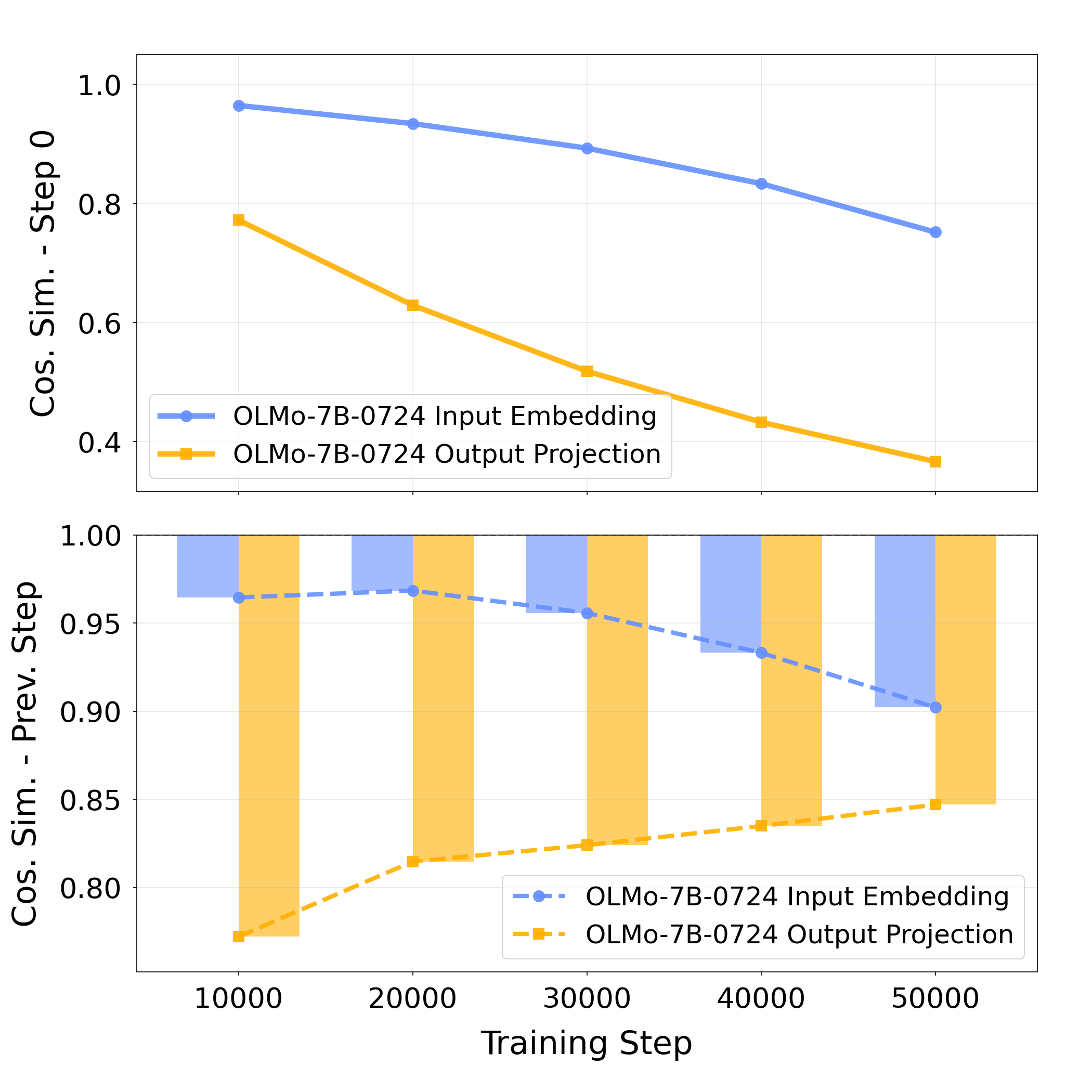}
    \caption{Embedding matrix dynamics for OLMo-7B (untied). Top: cosine similarity to initial embeddings (step 0), measuring cumulative drift. Bottom: cosine similarity between consecutive checkpoints, measuring per-step change rate. 
    }
    \label{fig:embedding-olmo7b}
\end{figure}

Figure~\ref{fig:embedding-olmo7b} shows OLMo-7B (untied) dynamics. The top panel (similarity vs initial) shows the cumulative effect: by step 50,000, the output matrix drifts to cosine similarity of approximately 0.38 with initialization, while input only drifts to 0.75. The bottom panel (consecutive checkpoints) reveals that output embeddings change much more rapidly than input embeddings, particularly in the first 10,000 steps where output similarity drops to approximately 0.78 while input remains above 0.96. This shows that the output embedding matrix undergoes significantly larger updates than the input embedding matrix in the early stages of training.

\begin{figure}[htbp]
    \centering
    \includegraphics[width=\columnwidth]{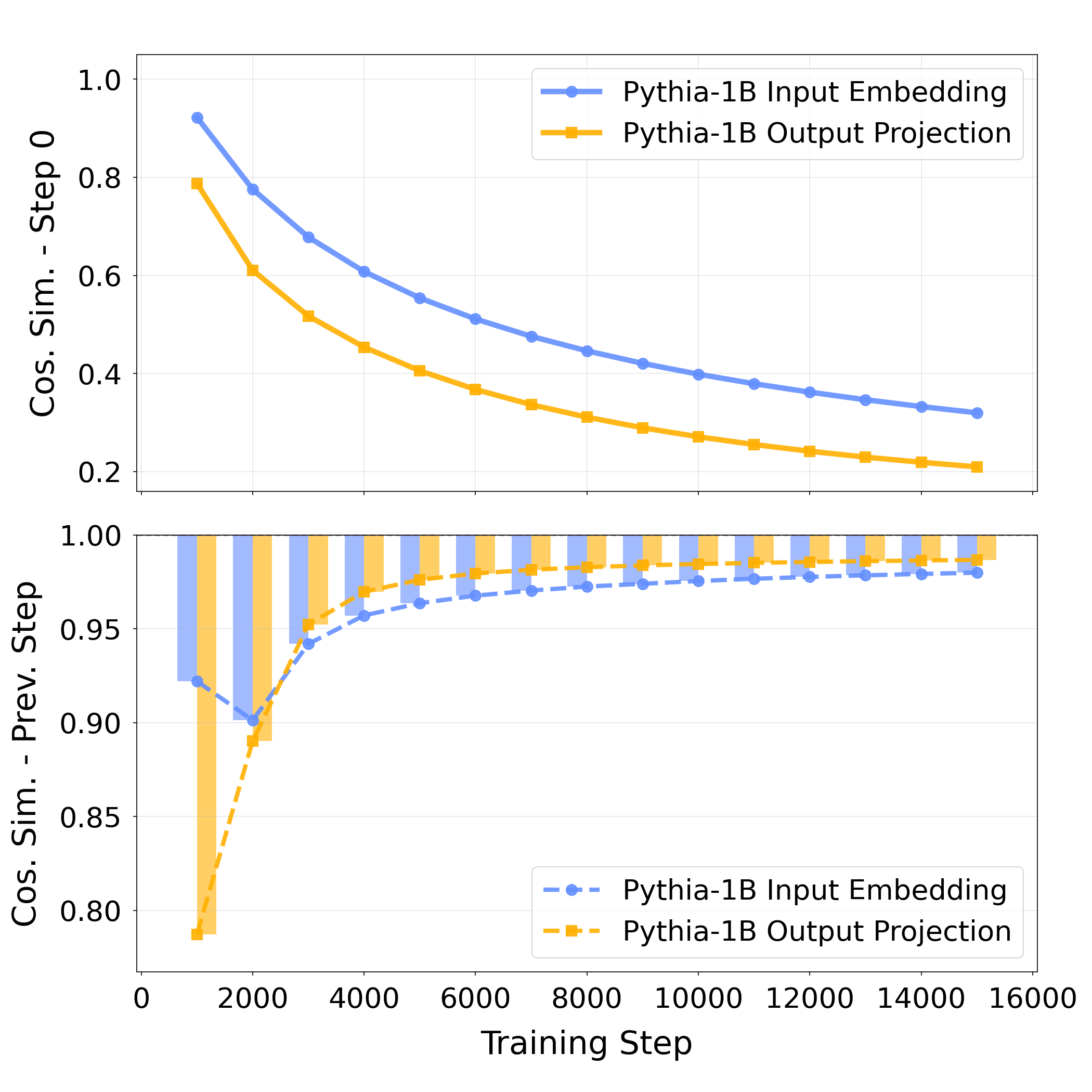}
    \caption{Embedding matrix dynamics for Pythia-1B (untied). Top: cosine similarity to initial embeddings (step 0), measuring cumulative drift. Bottom: cosine similarity between consecutive checkpoints, measuring per-step change rate. 
    }
    \label{fig:embedding-changes-pythia}
\end{figure}

Figure~\ref{fig:embedding-changes-pythia} shows Pythia-1B (untied). The top panel shows that output embeddings drift dramatically from initialization in the very first checkpoint, dropping to cosine similarity of approximately 0.79 (21\% change) while input only drops to 0.92 (8\% change). The bottom panel (per-step change rate) confirms this explosive early update: output reaches as low as 0.79 cosine similarity between the first two checkpoints, indicating massive restructuring, while input remains relatively stable. Both matrices then stabilize to similar per-step change rates after approximately step 4000. By the end of training (top panel), output has drifted to cosine similarity of 0.21 with initialization, while input has only drifted to 0.32.

The consistent pattern across OLMo-1B-0724 (Section~\ref{subsec:dominance-methodology}), OLMo-7B, and Pythia-1B establishes that output embeddings undergo substantially larger updates than input embeddings during early training, even when matrices are untied and receive gradients independently. This reflects the inherent structure of the training objective where output-layer gradients are naturally stronger than input-layer gradients.

\section{Additional Results --- Gradient Scaling}
\label{app:appendix-gradient-scaling}

Section~\ref{sec:gradient-intervention} presented gradient scaling results at step 10,000 for tied OLMo-1B models trained with input gradient scaling factors of 1$\times$ (baseline) and 5$\times$. Here we present additional results at step 1000 with scaling factors of 2$\times$ and 10$\times$ to demonstrate the robustness of the intervention effect. The experimental setup remains identical: we train OLMo-1B with tied embeddings from scratch using the standard configuration, applying gradient hooks to multiply input-layer gradients by a constant factor before they accumulate into the shared embedding matrix. At checkpoints, we measure cosine similarity (after Procrustes alignment) between the tied model's embedding matrix and the input/output matrices from OLMo-1B-0724 (untied) at matched training steps.

\begin{table}[h]
\centering
\small
\begin{tabular}{lcc}
\toprule
\textbf{Model} & \textbf{vs Untied Input} & \textbf{vs Untied Output} \\
\midrule
Tied (no scaling) & 0.172 & 0.197 \\
Tied (input $\times$2) & 0.173 & 0.197 \\
Tied (input $\times$10) & 0.173 & 0.190 \\
\bottomrule
\end{tabular}
\caption{Cosine similarity after Procrustes alignment of tied OLMo-1B models trained with scaled input gradients at step 1000. Even early in training, higher input gradient scaling slightly increases alignment with the untied input matrix and decreases alignment with the untied output matrix.}
\label{tab:gradient-scaling-step1000}
\end{table}

Table~\ref{tab:gradient-scaling-step1000} shows alignment results at step 1000, much earlier in training than the step 10,000 results in Table~\ref{tab:gradient-scaling} (Section~\ref{sec:gradient-intervention}). Even at this early stage, the intervention has a measurable effect. With baseline training (no scaling), the tied matrix achieves 0.172 cosine similarity with untied input and 0.197 with untied output, already showing the expected output bias. With 2$\times$ input gradient scaling, alignment with untied input increases slightly to 0.173 while alignment with untied output remains at 0.197. With 10$\times$ scaling, input alignment remains at 0.173 while output alignment decreases to 0.190.

The effect size is modest at step 1000, reflecting that only 2B tokens have been processed. However, the directionality is already correct: amplifying input gradients shifts the matrix toward input structure and away from output structure. The gradient scaling results across two checkpoints (step 1000 and step 10,000) demonstrate that gradient balance causally determines embedding structure, and this effect accumulates over training. The modest effect sizes reflect a fundamental constraint: improving input compatibility necessarily reduces output compatibility, confirming that weight tying forces a compromise between distinct representational demands.

\paragraph{Downstream evaluation.} A natural follow-up is whether the structural shift from gradient scaling translates into downstream performance gains. We evaluate the baseline tied model and the $5\times$-scaled model using lm-evaluation-harness \citep{eval-harness} on WikiText-2 \citep{merity2017pointer}, PIQA \citep{bisk2020piqa}, HellaSwag \citep{zellers2019hellaswag}, Winogrande \citep{sakaguchi2020winogrande}, ARC \citep{clark2018arc}, BoolQ \citep{clark2019boolq}, OpenBookQA \citep{mihaylov2018openbookqa}, and BLiMP \citep{warstadt2020blimp}, both at step 10,000.

\begin{table}[h]
\centering
\small
\begin{tabular}{lcc}
\toprule
\textbf{Benchmark} & \textbf{Tied} & \textbf{Tied (input $\times$5)} \\
\midrule
WikiText-2 PPL $\downarrow$ & \textbf{35.71} & 36.64 \\
PIQA & 0.620 & \textbf{0.631} \\
HellaSwag & \textbf{0.329} & 0.326 \\
Winogrande & \textbf{0.514} & 0.509 \\
ARC-Easy & 0.396 & \textbf{0.402} \\
ARC-Challenge & 0.233 & \textbf{0.241} \\
BoolQ & \textbf{0.576} & 0.548 \\
OpenBookQA & 0.256 & \textbf{0.278} \\
BLiMP & 0.755 & \textbf{0.757} \\
\bottomrule
\end{tabular}
\caption{Downstream evaluation of tied OLMo-1B models with and without input gradient scaling at step 10K (20B tokens). Higher is better for all benchmarks except WikiText-2 (perplexity, lower is better). No consistent performance difference is observed.}
\label{tab:downstream-scaling}
\end{table}

Table~\ref{tab:downstream-scaling} shows that although gradient scaling measurably shifts embedding structure (Table~\ref{tab:gradient-scaling}), this does not translate into consistent downstream improvements. Perplexity slightly worsens (35.71 $\rightarrow$ 36.64), consistent with the trade-off identified in Section~\ref{sec:gradient-intervention}: shifting the shared matrix toward input-like structure reduces its effectiveness as an output projection. Downstream results are mixed---the scaled model gains on PIQA, ARC-Challenge, and OpenBookQA, but loses on BoolQ, HellaSwag, and Winogrande, with no clear overall winner. BLiMP, a linguistic minimal-pair benchmark that directly tests grammatical competence, shows the two models are essentially identical (75.5\% vs 75.7\%). A naive constant scaling factor rebalances gradient contributions but does not resolve the underlying compromise.

\section{Chung et al. XTREME Benchmark Results}
\label{app:appendix-chung}

\begin{table*}[h]
    \centering
    \small
    \begin{tabular}{lcccccccc}
    \toprule
    & \textbf{\# PT} & \textbf{\# FT} & \textbf{XNLI} & \textbf{NER} & \textbf{PAWS-X} & \textbf{XQuAD} & \textbf{TyDi} & \textbf{Avg} \\
    & \textbf{params} & \textbf{params} & Acc & F1 & Acc & EM/F1 & EM/F1 & \\
    \midrule
    Coupled & 177M & 177M & 70.7 & \textbf{69.2} & \textbf{85.3} & 46.2/63.2 & 40.7/56.7 & 62.3 \\
    Decoupled & 269M & 177M & \textbf{71.3} & 68.9 & 85.0 & \textbf{46.9/63.8} & \textbf{42.8/58.1} & \textbf{62.7} \\
    \bottomrule
    \end{tabular}
    \caption{Effect of decoupling input and output embedding matrices on XTREME benchmark \citep{hu2020xtreme} performance \citep{chung2020rethinking}. PT: Pre-training. FT: Fine-tuning. Decoupling adds parameters during pre-training but improves average downstream performance.}
    \label{tab:chung-results}
    \end{table*}

Section~\ref{subsec:weight-tying-lms} references Chung et al. (2020) as key prior work demonstrating that weight tying can harm performance in transformer-based models. Table~\ref{tab:chung-results} displays their central finding from experiments on multilingual BERT-style models. They trained two configurations: coupled (tied embeddings, 177M parameters for both pre-training and fine-tuning) and decoupled (untied embeddings, 269M parameters during pre-training but pruned to 177M for fine-tuning to ensure fair comparison at evaluation time). Both models were pre-trained on multilingual data and evaluated on the XTREME benchmark, which tests cross-lingual transfer across diverse tasks. 

The decoupled model achieves 62.7\% average performance compared to 62.3\% for the coupled model. While the overall gain is modest (0.4 percentage points), improvements appear consistently across most tasks. XNLI (cross-lingual natural language inference) improves by 0.6 points (71.3 vs 70.7). The largest gains appear in question-answering tasks: XQuAD improves by 0.7/0.6 EM/F1 points (46.9/63.8 vs 46.2/63.2) and TyDi QA by 2.1/1.4 EM/F1 points (42.8/58.1 vs 40.7/56.7). These tasks may particularly benefit from improved input representations that support compositional reasoning.

Chung et al. (2020) established that untying helps but did not explain why. Our work provides the mechanistic explanation: decoupling allows input embeddings to specialize for semantic representation rather than being forced into an output-optimized space shaped by dominant output-layer gradients. Our tuned lens analysis (Section~\ref{subsec:tuned-lens-analysis}, Appendix~\ref{app:appendix-tuned-lens}) connects directly to these empirical results: elevated KL divergence in early layers of tied models indicates less effective compositional representations, which would particularly impact deep semantic understanding required for question answering.

\section{License Information}
\label{app:licenses}

GPT-Neo is released under an MIT license and Pythia, OLMo 3, and Qwen3 are all released under Apache 2.0 licenses. The Dolma dataset is released under an ODC-by license. All of these licenses permit use for scientific research and our use of these artifacts is consistent with their licenses.


\section{AI Assistants in Research or Writing --- Disclosure}
\label{app:ai-disclosure}

We used AI writing assistants (Claude, Cursor) during the preparation of this manuscript for editing and proofreading text, formatting LaTeX code, and implementing experiments. All scientific content, experimental design, analysis, and conclusions are the sole responsibility of the authors.

\end{document}